\documentclass[review]{elsarticle}

\usepackage{multirow}
\usepackage{amssymb}
\usepackage{amsmath}
\usepackage{lineno}
\usepackage{subcaption}
\usepackage{graphicx}
\usepackage{url}

\begin{document}

\title{Deep neural networks with Fisher vector encoding for medical image classification}

\author[1]{Lyra, Lucas O.}
\ead{lucas.oliveira.lyra@alumni.usp.br}
\author[1]{Fabris, Antonio E.}
\ead{aef@ime.usp.br}
\author[2]{Florindo, Joao B.}
\ead{florindo@unicamp.br}

\affiliation[1]{organization={Institute of Mathematics and Statistics of the University of Sao Paulo},
            addressline={Rua do Matao, 1010}, 
            city={Sao Paulo},
            postcode={05508-090}, 
            state={Sao Paulo},
            country={Brazil}}

\affiliation[2]{organization={Institute of Mathematics, Statistics and Scientific Computing of the University of Campinas},            
            addressline={Rua Sergio Buarque de Holanda, 651}, 
            city={Campinas},
            postcode={13083-859}, 
            state={Sao Paulo},
            country={Brazil}}

\begin{abstract}
Orderless encoding methods have shown to improve Convolutional Neural Networks (CNNs) for image classification in the context of limited availability of data. Additionally, hybrid CNN + Vision Transformers (ViT) models have been recently proposed to address CNN locality bias issues. These models outperformed CNN-only approaches. Despite that, the integration of such hybrid models with more elaborated feature representation can be highly beneficial and remains large unexplored in the literature. In this context, we propose the introduction of an orderless encoding method, Fisher Vectors, to hybrid CNN + ViT architectures, aiming at achieving a model suitable for both small and large datasets. Such enconding method relies on estimating a Gaussian Mixture Model (GMM) on image features. In large datasets, computational costs of the GMM estimation is a limiting factor for the application of Fisher Vectors. Thus, we propose a method to limit the growth of GMM estimation costs as we increase the size of the dataset. We explore the feasibility of our method in the context of medical image classification by appling it to MedMNIST (v2), Clean-CC-CCII and ISIC2018. This collection of datasets contains a wide variety of data scales and modalities. We outperform benchmark results in all MedMNIST (v2) datasets and obtain literature-competitive results in Clean-CC-CCII and ISIC2018.
\end{abstract}


\maketitle

\section{Introduction}

Convolutional Neural Networks (CNNs) have become the standard in image recognition in recent years. The mass adoption of CNN-based solutions in image classification and segmentation is attributable to the automatic feature extraction and the efficiency of its weight sharing \cite{yamashita2018convolutional}. 
Additionally, the hierarchical representation in CNNs allows learning different levels of features, which can be transferred to other visual recognition tasks \cite{gong2014multi,cimpoi2015deep}. 
However, CNNs require a fixed input size, which is a limiting factor for application in medical image classification, where images can vary largely in resolution \cite{manzari2023medvit}. 
This issue is normally addressed by pooling layers, such as average and max pooling. Nevertheless, pooling layers may lead to information loss \cite{li2022enhanced}.
In this context, the use of orderless pooling schemes based on Fisher Vectors in association with deep CNN have been proposed. These approaches have shown to improve CNN performance in small datasets \cite{yu2017hybrid, lyra2024multilevel}. This is particularly useful in medicine field, where constructing a large-scale dataset requires a significant amount of resources. It is necessary that experts in the field manually annotate and verify the data, which is time-consuming. \cite{manzari2023medvit}

An additional characteristic of CNNs is the locality bias \cite{yamashita2018convolutional}, which can limit their performance in image classification tasks as global information plays an important role in this case \cite{wang2018non}.
In recent years, this issue has been addressed through self-attention mechanisms, more specifically, Vision Transformers (ViT) \cite{dosovitskiy2020image}. 
However, ViT models require large amounts of data for training and are quadratically dependent on image resolution.
This makes such models unsuitable for direct application in medical field given the high resolution of images and possible limited data availability. In this sense, there is an increasing number of works associating ViT with CNN, creating hybrid models to address issues inherent to ViT \cite{,manzari2023medvit,li2022efficientformer,graham2021levit}.

Nevertheless, depite the promising results achieved by hybrid CNN+ViT models in multiple domains, there is still potential in exploring more sophisticated feature representation techniques within these architectures, especially focusing on good generalization in areas where data aquisition is expensive, as in medicine.

Based on this context, in this work, we propose to associate hybrid CNN + ViT models with Fisher Vector encoding. Our goal is to achieve a model that is suitable for various dataset sizes. 
While Fisher Vector encoding have good performance on small datasets \cite{lyra2024multilevel}, its application on large datasets remains a challenge. Such a method relies on estimating a Gaussian Mixture Model (GMM), which can be very memory demanding in big datasets. 
This is partially addreassed in \cite{lyra2024multilevel} by downsampling local features extracted from later convolutional layers, but such an approach may lose relevant information in the process. 
These issues are addressed in this work by applying a lossless approach for concatenating local features and proposing a method for subsampling the training dataset before estimating the GMM.
We show the robustness of our method by estimating GMMs for various subset sizes and using Kullback-Leibler (KL) Divergence to calculate their similarity. 
We evaluate the viability of our model on MedMNIST datasets \cite{yang2023medmnist}, where it outperforms benchmark results.
Finally, we evaluate our model in two datasets with more realistic image sizes: Clean-CC-CCII \cite{he2020benchmarking} and ISIC2018 \cite{tschandl2018ham10000,codella2018skin}. Our model show competitive results with the literature, outperforming recent models in Clean-CC-CCII.

In summary, the main contribuitions of this manuscript are:

\begin{itemize}
    \item We propose a GMM estimation method that consists of subsampling the training set based on image entropy. Experiments show that this methodology is robust and allows significant reduction in the number of samples used to train the GMM.
    \item A model for medical image classification is introduced, using a combination of hybrid CNN + ViT architectures and Fisher Vectors.
    \item A lossless approach was developed to concatenate features from multiple stages of the Transformer-based backbone, addressing issues reported in \cite{lyra2024multilevel}.
    \item Additionally, we have expanded the use of Fisher Vectors with neural networks to 3D image datasets, developing a methodology to retrieve knowledge from 2D domain.
\end{itemize}

Section~\ref{sec:related} provides an overview of the literature related to our study. 
In Section~\ref{sec:background}, we describe the theoretical basis required to present the proposed method. 
We detail, in Section~\ref{sec:proposed}, the proposed method for medical image classification. 
Section~\ref{sec:experiments} explains the experimental setup, describing the datasets used and implementation details.
In Section~\ref{sec:results}, we show and interpret the results of our experiments. Finally, in Section~\ref{sec:conclusions} we conclude our work with a general discussion and a proposal for future work.

\section{Related works}
\label{sec:related}

In this section we briefly explain some works that are related to this study. We start by reviewing the literature on Vision Transformers, followed by orderless encoding methods and medical image classification.

\subsection{Vision Transformers}

Following the significant progress of Transformer architectures in Natural Language Processing (NLP), ViT have been proposed for computer vision tasks such as image classification \cite{dosovitskiy2020image} and segmentation \cite{zheng2021rethinking}.
Standard ViT consists of dividing image into small patches and then applying transformer layers to them. Several works have been proposed to improve ViT performance. 
One example is the Pyramid Vision Transformers (PVT) model \cite{wang2021pyramid}, that attempts to reduce computational costs in high-resolution images by designing a progressive shrinking pyramid of transformers. 
Another example is the Swin Transformer model \cite{liu2021swin}, which addresses computational complexity of ViT with shifted window scheme and limiting the application of transformers to non-overlapping windows. 
Finally, the need to compute dot-product attention is eliminated in the Vision Outlooker (VOLO) architecture \cite{yuan2022volo}, that contains a new and light-weight attention mechanism.

Furthermore, hybrid models that combine CNN and ViT have been proposed to address issues in pure CNN or ViT-only architectures. 
LeViT \cite{graham2021levit} replaces patch embedding with a sequence of convolutional layers and also removes classification token in order to use average pooling. The model improves trade-off between accuracy and efficiency when compared to CNNs and ViT-based models. 
EfficientFormer \cite{li2022efficientformer}, besides replacing patch embedding with convolutional layers, includes the new proposed Meta Blocks, which removes the need of reshape operations. The model reduces latency in mobile devices when compared to other Transformer-based models.
MIT-EfficientViT \cite{cai2023efficientvit} replaces softmax self-attention with GPU-friendly operations to reduce computational costs of Transformers. It also includes a multi-scale linear attention module to address issues that arrise with the usage of the proposed self-attention mechanism.

\subsection{Ordeless encoding}

In the domain of visual texture classification, given the limited availability of data and the domain shift from ImageNet, numerous works have been proposed to take advantage of pre-trained weights from CNN. Most works use orderless encoding, as this is most suitable for this particular field. Cimpoi et. al. \cite{cimpoi2015deep} propose using Fisher Vectors to encode features extracted from the last convolutional layer of a VGG architecture \cite{simonyan2014very}. 
Lyra et. al. \cite{lyra2024multilevel} follow a similar approach, developing a novel strategy for extracting features from multiple convolutional layers. They apply the method in different CNN architectures and obtain improvements in accuracy when comparing to CNN-only approach. 

Besides Fisher Vector, other orderless encoding methods have been explored in literature. Chen et. al. \cite{chen2021deep} also extract features from multiple convolutional layers, however, the feature maps are concatenated by upsampling the lower resolution ones with bilinear interpolation. They measure statistical self-similarity of the resulting feature map using differential box counting method and compute soft histograms. Such histograms are concatenated with average pooling, resulting in the image descriptor. RADAM \cite{scabini2023radam} uses the same method for concatenating cross-layer feature maps, however the image descriptor is obtained from a proposed randomized AutoEncoder. In \cite{yang2022dfaen}, first- and second-order statistics from feature maps are extracted with a frequency attention mechanism and encoded using bilinear models. A combination of fractal average pooling and global average pooling is proposed in \cite{xu2021encoding} in order to obtain more robust descriptors. In \cite{florindo2023boff}, descriptors are computed with fuzzy equivalence measures over clustered local features.

\subsection{Medical Image Classification}

In the context of medical image classification, MedMNIST \cite{yang2023medmnist} have been proposed to offer a diverse group of datasets, with various sizes and modalities. Such work aimed to provide datasets on which generalization capability of models in medical image classification could be tested. The work also presents benckmark results using ResNet \cite{he2016deep} and AutoML approaches.

Many reseachers have used those datasets to evaluate their models. 
Zhang et al. \cite{zhang2023lc2r} propose the usage of a  longrange cross-residual mechanism to build a hybrid CNN + ViT. They show that such approach performs better than pure CNN and ViT models in MedMNIST. 
A hybrid CNN + ViT is also proposed in \cite{manzari2023medvit}, which is built using two proposed new blocks, an efficient convolutional block and a local transformer block. In \cite{liu2022feature} a hybrid ResNet + ViT model is constructed. In such model, multi-scale feature maps from ResNets are encoded using a transformer-based encoder.

Zheng et al. \cite{zheng2023complex} propose a method based on a complex-mixer network. They develop a pre-training framework to address issues of uncertainty and lack of information in the training dataset. 
In \cite{wang2024mednas} a novel neural architecture seach is developed in which they propose a new reduction cell seach space to handle the diversity of datasets. A generalist artificial inteligence model to handle both biomedical image and text is proposed in \cite{luo2023biomedgpt}. Finally, in \cite{khan2023medicat} a new training methodology is proposed to avoid overfitting when training ViT models with medical images.

\section{Background}
\label{sec:background}

In this section, we describe two core principles behind our methodology. The first is attention mechanism, which is is responsible here for feature extraction. The second is Fisher Vector encoding, which provides the feature vectors employed for classification.

\subsection{Attention mechanism}

Given an input vector $x\in\mathrm{R}^{N\times f}$ and three learnable projection matrices $W_Q \in\mathrm{R}^{f\times d}$, $W_K\in\mathrm{R}^{f\times d}$, and $W_V\in\mathrm{R}^{f\times d}$, we calculate the query, key and value matrices respectively by $Q=xW_Q$, $K=xW_K$, and $V=xW_V$.
The attention $A$ can be computed by:

\begin{equation}
    A(Q,K,V) = Sim(Q,K)V,
\end{equation}
\noindent
where $Sim(Q,K)$ is a similarity function between matrices $Q$ and $K$. 
In original softmax self-attention model \cite{vaswani2017attention}, we have
\begin{equation}
    Sim(Q,K) = \text{softmax}\left(\frac{QK^{T}}{\sqrt{d}}\right).
\end{equation}

In this work, local features are generated using the model proposed in \cite{cai2023efficientvit}, which uses a ReLU linear attention. The similarity function is this case is given by 
\begin{equation}
    Sim(Q,K) = \frac{\text{ReLU}(Q)\text{ReLU}(K)^{T}}{\sum_{j=1}^{N}\text{ReLU}(Q)\text{ReLU}(K_j)^{T}},
\end{equation}
where $K_j$ is the $j$-th row of $K$.

\subsection{Kullback-Leibler Divergence}

The KL Divergence is a measure of similarity between two probability density functions. It is also known as \emph{relative entropy}. Given two density functions $f$ and $g$, the KL Divergence is defined as:

\begin{equation}
    \label{eq:kl_divergence}
    D(f||g) = \int_{-\infty}^{\infty} f(x)\log\left(\frac{f(x)}{g(x)}\right)dx.
\end{equation}

Equation~\ref{eq:kl_divergence} does not have a closed-form expression when $f$ and $g$ are GMMs. In this case, we can only estimate it numerically. There are various ways of doing that, and the most accurate approach is Monte Carlo sampling \cite{hershey2007approximating}. In this case, we have

\begin{equation}
    \label{eq:mc_estimation}
    D_{MC}(f||g) = \frac{1}{n} \sum_{i=1}^{n} \log\left(\frac{f(x_i)}{g(x_i)}\right),
\end{equation}
where $x_i$ is a sample drawn from the GMM $f$ and $n$ is the number of samples. Equation~\ref{eq:mc_estimation} converges to $D(f||g)$ as $n\to\infty$.

\subsection{Fisher Vector}

Fisher Vectors are usually calculated employing a mixture of Gaussian distributions to model the probability of the ocurrence of a local feature.
We assume that this probability can be written as
\begin{equation}
    p(\mathbf{x}) = \sum_{k=1}^{K}w_k\mathcal{N}(\mathbf{x}|\boldsymbol{\mu}_k,\boldsymbol{\Sigma}_k),
\end{equation}
where $\mathbf{x}$ is a local feature, $K$ is the number of Gaussian in the mixture, $w_k$, $\boldsymbol{\mu}_k$ and $\boldsymbol{\Sigma}_k$ are respectively the weight, mean vector, and covariance matrix of the $k$-th distribution.

Given the observation of the local feature $\mathbf{x}$, we can estimate the probability of this local feature being sampled from the $k$-th distribuition in the mixture as
\begin{equation}
\gamma_k(\mathbf{x}) = \frac{w_k \mathcal{N}(\mathbf{x}|\boldsymbol{\mu}_k,\boldsymbol{\Sigma}_k)}{\sum_{i=1}^{K} w_i \mathcal{N}(\mathbf{x}|\boldsymbol{\mu}_i,\boldsymbol{\Sigma}_i)}.
\end{equation}

We also assume that the covariance matrices are diagonal. We denote by $\boldsymbol{\sigma}_{k}^2$ the vector of diagonal elements of $\boldsymbol{\Sigma}_k$. In this context, each component of the Fisher Vector can be calculated by the following three equations:
\begin{align}
    \label{eq:fisher1}
    \mathcal{G}_{w_k^d}^{X} &= \frac{1}{T\sqrt{w_k}}\sum_{t=1}^{T} 
    \left( \gamma_k(\mathbf{x}_t)- w_k\right), \\
    \label{eq:fisher2}
    \mathcal{G}_{\mu_k^d}^{X} &= \frac{1}{T\sqrt{w_k}}\sum_{t=1}^{T} \gamma_k(\mathbf{x}_t)\left(\frac{x_t^d-\mu_k^d}{\sigma_k^d}\right), \\
    \label{eq:fisher3}
    \mathcal{G}_{\sigma_k^d}^{X} &= \frac{1}{T\sqrt{2w_k}}\sum_{t=1}^{T} \gamma_k(\mathbf{x}_t)\left[\frac{(x_t^d-\mu_k^d)^2}{(\sigma_k^d)^2}-1\right].
\end{align}
In these equations, $d$ denotes the $d$-th element of the respective vector and $T$ denotes the number of local features. Additional information on Fisher Vectors can be found in \cite{sanchez2013image}.

\section{Proposed method}
\label{sec:proposed}

\begin{figure}
    \centering
    \includegraphics[width=\textwidth]{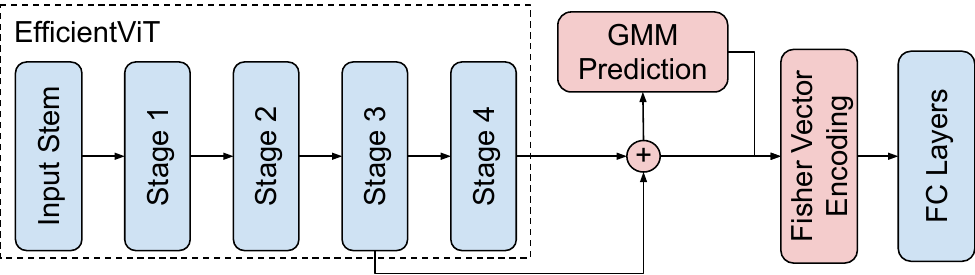}
    \caption{Proposed scheme for using Fisher Vector encoding associated with MIT-EfficientViT architecture. In this representation, local features are extracted from two stages. However, different numbers of stages can be used. We apply layer normalization to the first FC layer as performed in MIT-EfficientViT. When fine-tuning is applied, weights from stages 1 to 4 are allowed to vary. Classifier training affects exclusively weights from FC layers. }
    \label{fig:layout}
\end{figure}

We propose the use of MIT-EfficientViT \cite{cai2023efficientvit}, a multi-scale vision transformer model, to extract local features for Fisher Vector encoding.  
This is achieved by changing the model's head. The head is composed by Mobile inverted Bottleneck Convolution (MBConv) and two Fully-Connected (FC) layers. We replace the MBConv layer with Fisher Vector enconding. 
The encoding is followed by the normalization proposed in \cite{perronnin2010improving}.
We maintain the model's classifier, modifying the number of input features in the first FC layer to match the size of the Fisher Vector. The proposed architecture is presented in Figure~\ref{fig:layout}.

MIT-EfficientViT is an architecture proposed for 2D images. In order to enable it to handle 3D images, we replace all the convolutional layers with 3D ones. This includes depth-wise convolutions (DWConv) from the Multi-Scale Linear Attention mechanism proposed in the MIT-EfficientViT model.

We initalize the model with pre-trained weights from ImageNet provided by the Timm library \cite{rw2019timm}. Such weights provide local features with decent generalization ability \cite{li2022domain}. However, we want to generate application-specific high-level features. This is achieved by fine-tuning the backbone before using it to extract local features for Fisher Vector encoding. Fine-tuning is also useful in the case of 3D images, because there are no pre-trained weights for this kind of applicaiton. In order to initialize the model in this later case, we repeat 2D pre-trained weights over one axis.

Local features are extracted from 1 to 3 stages of the backbone, prioritizing the later ones. This means that, when one layer is used, we extract local features from Stage 4 exclusively. Notably, such priority ensures that attention mechanism will always be present in local feature extraction.

Each stage contains local features in a different dimensional space. As such, concatenating local features in a single set requires projecting them into the same dimensional space.
The approach taken in \cite{lyra2024multilevel} to solve this issue is downsampling local features that belong to the higher dimensional space, but such solution leads to information loss. If we consider upsampling features from the lower dimensional space, information loss is avoided, however, it increases the computational cost of GMM training.

Hence, we propose a lossless method of concatenation of features that does not require upsampling. 
Let $n$ be the number of stages for feature extraction. Let $d_i$ be the dimension of local features from the Stage $n-i$. Our approach consists in fiding a dimension $d$ such that 
\begin{equation}
    d_i = c_i d \quad \forall i \in \{0, 1,\cdots,n-1\},
\end{equation}
where $c_i\in\mathbb{N}$. Each local feature from stage $i$ is splitted into $c_i$ local features with dimension $d$. Finally, splitted local features in a single set are concatenated. In this study, we choose
\begin{equation}
    d = \mathrm{gdc}(d_1, d_2, \cdots, d_n),
\end{equation}
\noindent
where $\mathrm{gdc}$ stands for Greatest Common Divisor.
In the particular case of MIT-EfficientViT, $d=384$ when $n=1$, $d=192$ when $n=2$, and $d=96$ when $n=3$, were empirically found to work appropriately.

Finally, we propose to limit resources used in GMM estimation by setting a limit to the number of samples that can be used for this purpose.
Let $\mathbf{X}$ denote the set of all brightness values observed in an image. We divide $\mathbf{X}$ into $k$ evenly spaced bins by computing values $X_{i}$, $i\geq 0$, as follows:
\begin{equation}
    X_{i} = \frac{i}{k}\left(\max(\mathbf{X})-\min(\mathbf{X})\right) + \min(\mathbf{X}),
\end{equation}
where $\max(\cdot)$ and $\min(\cdot)$ denote, respectively, the maximum and minimum values observed in a given set.

Additionally, let $\mathbb{I}(e)$ denote the binary indicator function, which is defined as
\begin{equation}
\label{eq:indicator}
    \mathbb{I}(e) \triangleq 
    \begin{cases} 
      0 & \text{if $e$ is false,} \\
      1 & \text{if $e$ is true}.
   \end{cases}
\end{equation}
We estimate the probability density function $p$ of observing a given brightness value $x$ as 
\begin{equation}
    \label{eq:probability}
    p(x) = \frac{1}{|\mathbf{X}|} \sum_{i=1}^{k} \mathbb{I}(X_{i-1} < x \leq X_{i}),
\end{equation}
where $|\cdot|$ denotes the cardinality of a given set.
Using the probability estimation in Equation~\ref{eq:probability} we evalute the Shannon entropy of an image $H(\mathbf{X})$ by
\begin{equation}
    H(\mathbf{X}) \triangleq -\sum_{x\in\mathbf{X}} p(x) \log p(x).
\end{equation}
Samples for GMM estimation are withdrawn, without replacement, from the training set from highest to lowest entropy value up to a limit of samples $n$.

\section{Experimental setup}
\label{sec:experiments}

In this section we provide detailed information about our experimental setup. We describe the datasets used, model architecture, optimizers, and parameters used for training and other relevant information to reproduce our results.

\subsection{Datasets}

We evalaute our model in all the MedMNIST (v2) \cite{yang2023medmnist} datasets, as these datasets are low-weight and designed for model validation. We proceeed to apply our model in two datasets with more realistic image sizes: ISIC2018 \cite{tschandl2018ham10000,codella2018skin} and the Clean-CC-CCII \cite{he2020benchmarking}. In the following subsections, we describe each one of the datasets.

\begin{figure}
    \centering
    \includegraphics[width=\textwidth]{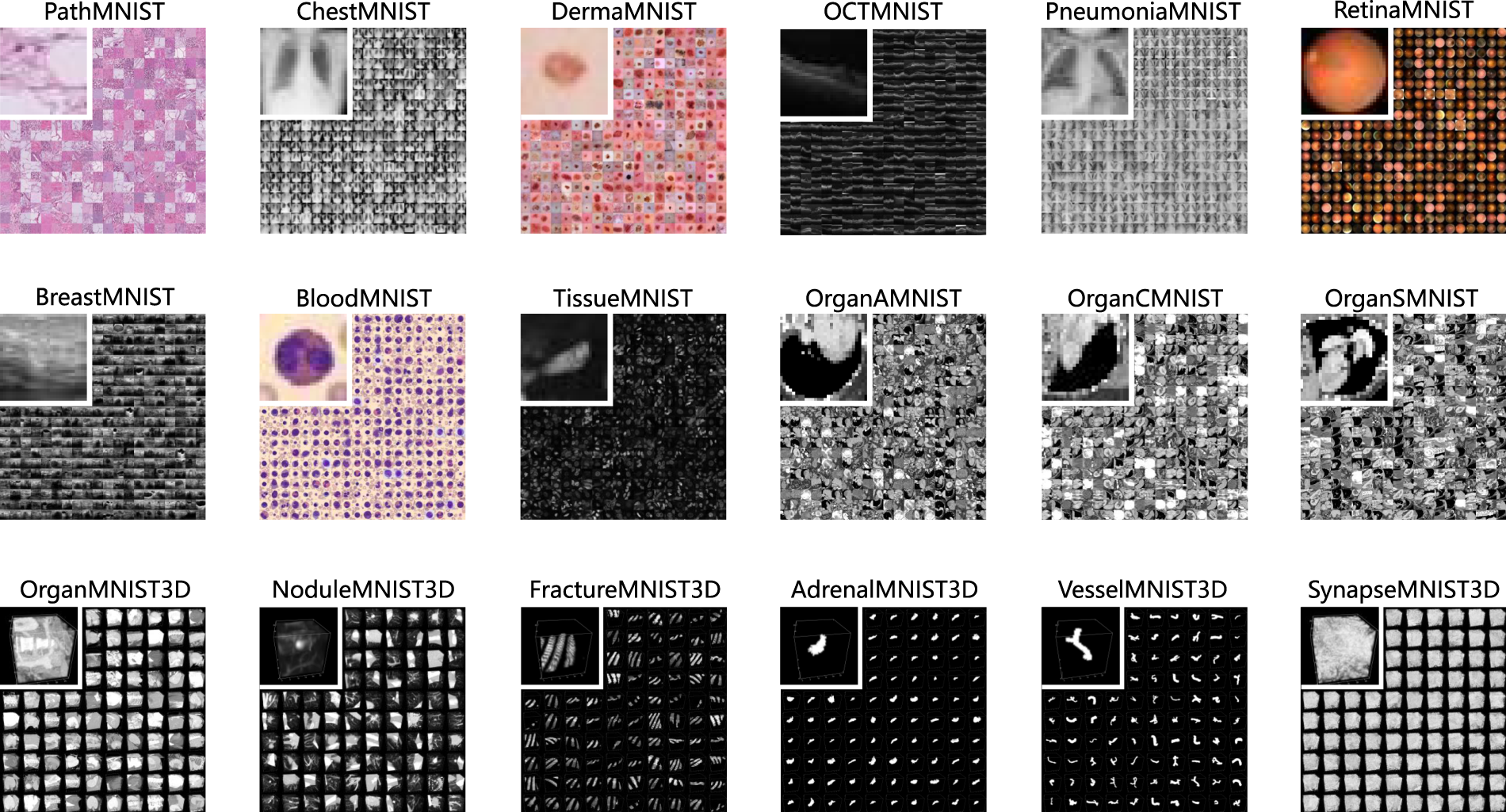}
    \caption{MedMNIST (v2) collection. It consists of twelve 2D datasets and six 3D datasets. All images have width and height (and depth when applicable) equal to 28 pixels. This image was retrieved from \cite{yang2023medmnist}.}
    \label{fig:medmnist}
\end{figure}

The MedMNIST (v2) is collection of 9 datasets of gray-scale 2D images, 3 datasets of colored 2D images, and 6 datasets of single-channel 3D images. Tasks include binary, multi-class and multi-label classification. All images are sized 28x28 in 2D case and 28x28x28 in 3D case. The dataset is exemplified in Figure~\ref{fig:medmnist}, which was obtained from the original article.

The three colored 2D datasets are RetinaMNIST, DermaMNIST, and BloodMNIST. RetinaMNIST is based on the DeepDRiD challenge \cite{liu2022deepdrid} and consists of 1,600 retina fundus images divided into 5 classes. Those classes correspond to the diabetic retinopathy severity. DermaMNIST consists of 10,015 dermatoscopic images, which are categorized into seven different diseases. BloodMNIST consists of 17,092 microscopic blood cell images, each one belonging to one out of eight classes.

The datasets OrganAMNIST, OrganCMNIST, and OrganSMNIST consist of 2D slices of 3D computed tomography (CT) images based on the Liver Tumor Segmentation Benchmark \cite{bilic2023liver}. Planes used for slicing are axial, coronal, and sagittal, respectively. The task consists of identifing body organs, which are 11 in total. OrganAMNIST contains 58,850 images, OrganCMNIST contains 23,660, and OrganSMNIST contains 25,221.

ChestMNIST and PneumoniaMNIST consist of chest X-ray images and are based on the datasets proposed in \cite{wang2017chestx} and \cite{kermany2018identifying}, respectively. In the first case, there are 112,120 images and the task is multi-label classification of 14 possible diseases. The second case consists of 5,856 images and the task is the diagnostic of pneumonia.

The remaining 2D datasets are PathMNIST, TissueMNIST, OCTMNIST, and BreastMNIST. PathMNIST consists of identityfing one out of 9 types of tissues in 107,180 histological images related to colorectal cancer. TissueMNIST consists of 236,386 microscope images of kidney cortex cells, which are divided into 8 classes. OCTMNIST contains 109,309 optical coherence tomography images for retinal diseases, which are divided into 4 diagnostic categories. BreastMNIST contains 780 breast ultrasound images, which are categorized as normal, benign, or malignant.

NoduleMNIST3D and FractureMNIST3D both consist of chest CT scans. The first one contains 1,633 images divided into negative and positive cases of lung nodules. The second one contains 1,370 images and the task is classifying three cases of rib fractures: buckle, nondisplaced, and displaced.

VesselMNIST3D is a set of 3D models of entire brain vessels built from Magnetic Resonance Angiography (MRA) images. It contains 1,909 images divided into negative and positive cases of aneurysm. Synapse3D consists of 1,759 3D images acquired by multi-beam scanning electron microscope. The task is distinguishing between excitatory and inhibitory synapses.

AdrenalMNIST3D and OrganMNIST3D are both datasets of abdominal CTs. The later consists of 1,743 images and the task is classifying 11 body organs. The former contains 1,584 shape masks from adrenal glands and the task is identifying whether the adrenal gland is normal.

The Clean-CC-CCII dataset is a clean and segmented dataset build based on chest CT scans. It is divided into positive and negativa cases of COVID-19. It contains 425 scans used for training, 118 used for validation and 203 for testing. Images are presented in varying sized and shapes, so we proceed to crop images at the center to fix shape and resize then to 224x224 to fix size.

The ISIC2018 dataset consists of dermatoscopic images of common pigmented skin lesions. It is divided in seven classes: actinic keratoses and intraepithelial carcinoma (AKIEC), basal cell carcinoma (BCC), benign keratosis (BKL), dermatofibroma (DF), melanocytic nevi (NV), melanoma (MEL), vascular skin lesions (VSK). We use the training datasets that consists of 10015 images. We radomly split it utilizing 70\% of samples for training and the remaining for testing. We employ the same procedure used in Clean-CC-CCII to fix the size of the images.

\subsection{Implementation details}

In all experiments, we normalize images to $0.5$ mean and $0.5$ standard deviation. Unless otherwise specified, we resize image width and height of MedMNIST(v2) from 28 to 224. 
The interpolation method used is repeating the nearest value. We compare our model with results from \cite{yang2023medmnist,wang2024mednas,zhang2023lc2r,zheng2023complex}. 

We use the MIT-EfficientViT-B2 architecture \cite{cai2023efficientvit} as the backbone for feature extraction. Unless otherwise specified, we use the fine-tuned backbone. The optimizer used for fine-tuning and classifier training is chosen between Adam \cite{kingma2014adam} and RAdam \cite{liu2019variance} based on the validation set. The number of epochs is set to 20, with early stopping based on the validation set. The learning rate for backbone fine-tuning is set to $10^{-3}$. The learning rate for classifier training is set to $10^{-4}$. In all cases, a decay of $0.1$ is applied after epochs 10 and 15. We use cross-entropy loss as criterion for training.

In order to handle input images with 1 channel instead of 3, we reshape the first layer of the backbone and sum pre-trained weights from the original 3 channels. For managing 3D images, we replace 2D convolutional layers for 3D ones. In order to leverage pre-trained weights, we repeat 2D weights over one axis. The choice of axis is based on the validation set.

Dictionary learning from local features is performed by GMM with 16 components (kernels).
Unless otherwise specified, local features are extracted from two stages.
We limit the number of training samples in GMM to 5,000. We set regularization parameter to $10^{-4}$ multiplied by the maximum standard deviation of the local features. 

\subsection{Evaluation Metrics}

Our method was evaluated using Accuracy (ACC) and the Area Under ROC Curve (AUC), where ROC stands for Receiver Operating Characteristic. In the following paragraphs, we explain how ACC and AUC are computed in binary, multi-class and multi-label classification tasks.

Let the number of samples be denoted by $n$ and the number of possible labels be denoted by $k$. Given a sample of index $i$, we denote the true value of its $j$-th label by $Y_{i}^{j}$ and the respective prediction by $Z_{i}^{j}$.
For all classification tasks, ACC is given by
\begin{equation}
    \text{ACC} = \frac{1}{n}\sum_{i=1}^{n} \frac{1}{k}\sum_{j=1}^{k} \mathbb{I}(Y^{j}_i = Z^{j}_i),
\end{equation}
where $\mathbb{I}$ is the indicator function defined in Equation~\ref{eq:indicator}.

Consider both the binary and multi-label classification tasks. Let $D_{0,j}$ and $D_{1,j}$ respectively denote the set of negative and positive examples for the $j$-th label. Given a probability prediction function $f_j$ learned by a classifier for the $j$-th label, AUC is computed by
\begin{equation}
\label{eq:auc}
    \text{AUC} = \frac{1}{k}\sum_{j=1}^{k}\frac{1}{|D_{0,j}|}\sum_{t_0\in D_{0,j}}\frac{1}{|D_{1,j}|}\sum_{t_1\in D_{1,j}}\mathbb{I}\left(f_j(t_0)<f_j(t_1)\right),
\end{equation}
where $|\cdot |$ denotes the number of elements in a given set.

When evaluating AUC for multi-class classification tasks, we adopt the \emph{one-vs-rest} strategy. For each possible class $c$, Equation~\ref{eq:auc} is evaluated considering this class as a positive example and the remaining as negative examples. The resulting AUC is given by the simple average of each evaluation.

\section{Results and Discussion}
\label{sec:results}

\begin{figure}
\centering
\begin{subfigure}{.48\textwidth}
  \centering
  \includegraphics[width=\linewidth]{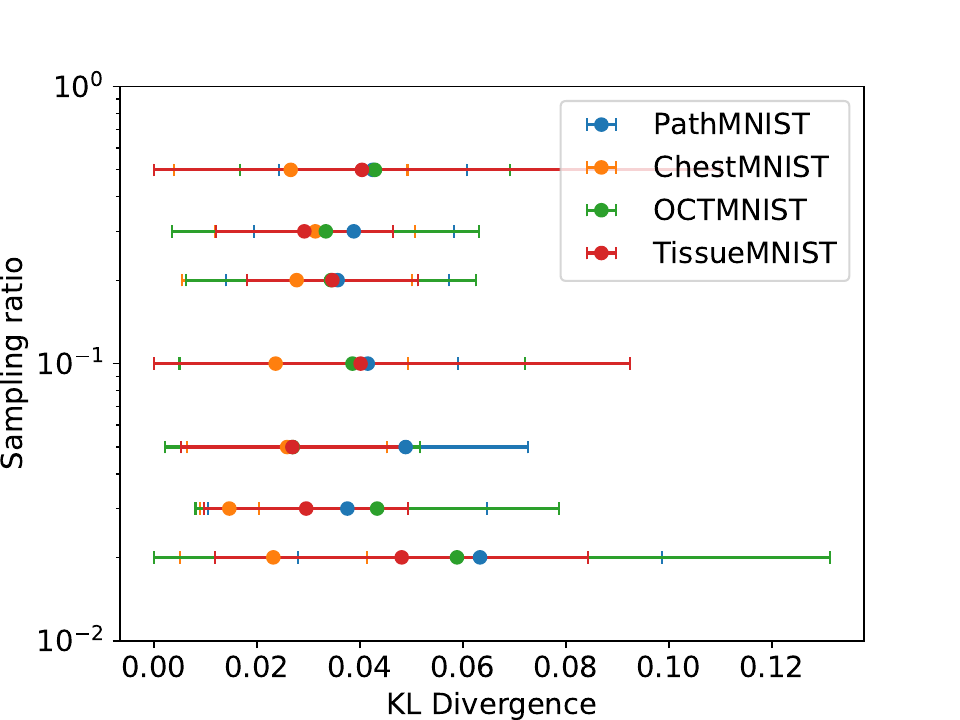}
  \caption{KL Divergence between the base case GMM and the GMMs estimated using a subset of the training dataset.}
  \label{fig:kl_divergence}
\end{subfigure}
\begin{subfigure}{.48\textwidth}
  \centering
  \includegraphics[width=\linewidth]{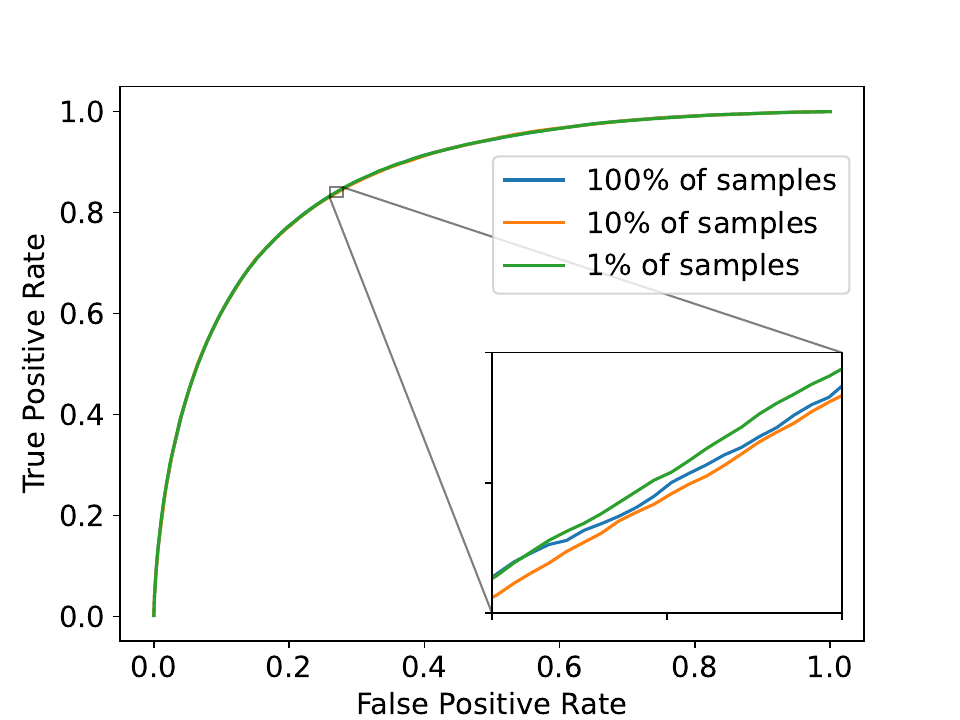}
  \caption{Macro-averaged One-vs-Rest ROC Curve for TissueMNIST}
  \label{fig:roc_curve}
\end{subfigure}
\caption{(a) KL Divergence mean and standard deviation between the base case GMM and the GMM estimated using a subset of the dataset. Sampling ratio stands for the ratio between the size of the subset and the size of the original training set. (b) Impact of sampling the training set on the ROC Curve.}
\label{fig:gmm_estimation}
\end{figure}

According to Lyra et. al. \cite{lyra2024multilevel}, the applicability of Fisher Vector encoding is restricted to small datasets given the scalability of GMM estimation.
Nevertheless, by using the approach proposed in this paper, we are capable of scaling GMM estimation to larger datasets. In Figure ~\ref{fig:gmm_estimation}, we evaluate how sampling a dataset can impact on GMM estimation. We select the four largest datasets, all of them containing more than 100,000 images. We use the pre-trained backbone and keep all images in their original dimensions. 

We first compute the base case GMM, which is the GMM estimated using the entire training set. 
We proceed to sampling the training set using our proposed approach. For each subset, we estimate a new GMM using the same parameters as the base case GMM.  
We compute the similarity between the later and the former using the KL Divergence. 
As there is no closed-form solution to compute the KL Divergence for GMMs, we estimate it using the Monte-Carlo method with 1 million samples \cite{hershey2007approximating}. This method was preferred over others because of its accuracy \cite{popovic2021measure}.
Given the randomness of GMM initialization, the procedure described in this paragraph is repeated 10 times, varying the initialization seed.

In Figure~\ref{fig:kl_divergence}, we show that a GMM estimated using $2\%$ of the training set is as similar to the base case GMM as a GMM estimated using half of the training set. 
This indicates that, up to a certain limit, the reduction of samples does not statistically significantly affect GMM estimation.
In Figure~\ref{fig:roc_curve}, we show, in the case of the larggest dataset, that reducing the number of samples does not impact the model's accuracy. This behavior can actually be verified for the other datasets. We chose this particular case because this dataset is most impacted by setting a fixed limit of samples for GMM estimation.

We provide a study on the impacts of backbone fine-tuning in ACC and AUC. This evaluation is done on a general perspective for 2D and 3D cases. 
We calculate the simple average of metrics when we use pre-trained weights and when we fine-tune the backbone.
Results are shown in Table~\ref{tab:finetune_mnist}. In the 3D case, as no pre-trained weights are available, we repeat 2D weights over one axis. Notably, the axis used is the same as in other experiments.
In both 2D and 3D cases, fine-tuning the backbone improved all metrics of evaluation of the model. Greater improvements, especially in AUC, can be observed in the 3D case. This is expected given the lack of pre-trained weights for this specific type of image.

\begin{table}
\centering
\caption{Average Accuracy and Average Area Under ROC Curve of the model with pre-trained and fine-tuned backbone. We show the results for varying number of stages used for feature extraction. Stages are chosen from later to earlier ones.}
\label{tab:finetune_mnist}
\resizebox{\linewidth}{!}{
\begin{tabular}{ l c | c c | c c }
\multirow{2}{*}{Datasets} & \multirow{2}{*}{Stages} & \multicolumn{2}{c|}{\textit{Pre-trained backbone}} & \multicolumn{2}{c}{\textit{Fine-tuned backbone}} \\
\cline{3-6}
 &  & Average ACC & Average AUC & Average ACC & Average AUC \\
 \hline
 MedMNIST2D & 1 & 0.809 & 0.921 & 0.849 & 0.935 \\ 
 MedMNIST2D & 2 & 0.826 & 0.929 & 0.853 & 0.939 \\ 
 MedMNIST2D & 3 & 0.825 & 0.928 & 0.849 & 0.938 \\ 
 \hline
 MedMNIST3D & 1 & 0.794 & 0.809 & 0.841 & 0.899 \\ 
 MedMNIST3D & 2 & 0.794 & 0.828 & 0.851 & 0.906 \\
 MedMNIST3D & 3 & 0.799 & 0.824 & 0.836 & 0.889 \\ 
\end{tabular}
}
\end{table}

In Table~\ref{tab:finetune_mnist}, we also show how the model is impacted by the choice of the number of stages to extract local features. 
One stage means features are extracted from the Stage 4. Two stages is represented in Figure~\ref{fig:layout}. Three stages means we also include local features from the Stage 2. 
When using the pre-trained backbone, chosing two or three stages for feature extraction yields similar results. However, the use of two stages outperforms other choices when fine-tuning is applied. This may indicate that, with fine-tuning, relevant information for classification is being captured in later layers. This causes the use of earlier layers to be detrimental. In fact, with fine-tuning, using exclusively the last stage is better, in terms of AUC and ACC, than retrieving information from stages as early as the Stage 2.

We compare our model with benckmark results that are suitable for both 2D and 3D cases. We use the configuration specified in Section~\ref{sec:experiments}.  Results are shown in Table~\ref{tab:benchmark_mnist}. All presented values were obtained from the original MedMNIST(v2) paper\cite{yang2023medmnist}.
In all cases, our model outperforms benchmark, with most outstanding results observed in RetinaMNIST, OrganMNIST3D, FractureMNIST3D and SynapseMNIST3D, where accuracy was improved by at least 5\%.  The datasets do not share similarities, as they vary in classification task and data modality, what indicates the model can be used in a large variaty of image classification tasks.


\begin{table}
    \centering
    \resizebox{\linewidth}{!}{
    \begin{tabular}{l | c c | c c | c c | c c | c c | c c }
        \hline
         \multirow{2}{*}{Methods} & \multicolumn{2}{c|}{\textit{PathMNIST}} & \multicolumn{2}{c|}{\textit{ChestMNIST}} & \multicolumn{2}{c|}{\textit{DermaMNIST}} & \multicolumn{2}{c|}{\textit{OCTMNIST}} & \multicolumn{2}{c|}{\textit{PneumoniaMNIST}} & \multicolumn{2}{c}{\textit{RetinaMNIST}} \\
         \cline{2-13}
          & AUC & ACC & AUC & ACC & AUC & ACC & AUC & ACC & AUC & ACC & AUC & ACC \\
          \hline
        ResNet-18 (28) & 0.983 & 0.907 & 0.768 & 0.947 & 0.917 & 0.735 & 0.943 & 0.743 & 0.944 & 0.854 & 0.717 & 0.524 \\
        ResNet-18 (224) & 0.989 & 0.909 & 0.773 & 0.947 & 0.920 & 0.754 & 0.958 & 0.763 & 0.956 & 0.864 & 0.710 & 0.493 \\
        ResNet-50 (28) & 0.990 & 0.911 & 0.769 & 0.947 & 0.913 & 0.735 & 0.952 & 0.762 & 0.948 & 0.854 & 0.726 & 0.528 \\
        ResNet-50 (224) & 0.989 & 0.892 & 0.773 & \textbf{0.948} & 0.912 & 0.731 & 0.958 & 0.776 & 0.962 & 0.884 & 0.716 & 0.511 \\
        auto-sklearn & 0.934 & 0.716 & 0.649 & 0.779 & 0.902 & 0.719 & 0.887 & 0.601 & 0.942 & 0.855 & 0.690 & 0.515 \\
        AutoKeras & 0.959 & 0.834 & 0.742 & 0.937 & 0.915 & 0.749 & 0.955 & 0.763 & 0.947 & 0.878 & 0.719 & 0.503 \\
         Ours & \textbf{0.991} & \textbf{0.917} & \textbf{0.792} & \textbf{0.948} & \textbf{0.934} & \textbf{0.781} & \textbf{0.964} & \textbf{0.786} & \textbf{0.975} & \textbf{0.897} & \textbf{0.764} & \textbf{0.568} \\  
         \hline\hline
         \multirow{2}{*}{Methods} & \multicolumn{2}{c|}{\textit{BreastMNIST}} & \multicolumn{2}{c|}{\textit{BloodMNIST}} & \multicolumn{2}{c|}{\textit{TissueMNIST}} & \multicolumn{2}{c|}{\textit{OrganAMNIST}} & \multicolumn{2}{c|}{\textit{OrganCMNIST}} & \multicolumn{2}{c}{\textit{OrganSMNIST}} \\
         \cline{2-13}
          & AUC & ACC & AUC & ACC & AUC & ACC & AUC & ACC & AUC & ACC & AUC & ACC \\
          \hline
        ResNet-18 (28) & 0.901 & 0.863 & 0.998 & 0.958 & 0.930 & 0.676 & 0.997 & 0.935 & 0.992 & 0.900 & 0.972 & 0.782 \\
        ResNet-18 (224) & 0.891 & 0.833 & 0.998 & 0.963 & 0.933 & 0.681 & 0.998 & 0.951 & 0.994 & 0.920 & 0.974 & 0.778 \\
        ResNet-50 (28) & 0.857 & 0.812 & 0.997 & 0.956 & 0.931 & 0.680 & 0.997 & 0.935 & 0.992 & 0.905 & 0.972 & 0.770 \\
        ResNet-50 (224) & 0.866 & 0.842 & 0.997 & 0.950 & 0.932 & 0.680 & 0.998 & 0.947 & 0.993 & 0.911 & 0.975 & 0.785 \\
        auto-sklearn & 0.836 & 0.803 & 0.984 & 0.878 & 0.828 & 0.532 & 0.963 & 0.762 & 0.976 & 0.829 & 0.945 & 0.672 \\
        AutoKeras & 0.871 & 0.831 & 0.998 & 0.961 & 0.941 & 0.703 & 0.994 & 0.905 & 0.990 & 0.879 & 0.974 & 0.813 \\
        Ours & \textbf{0.922} & \textbf{0.897} & \textbf{0.999} & \textbf{0.977} & \textbf{0.947} & \textbf{0.726} & \textbf{0.999} & \textbf{0.959} & \textbf{0.997} & \textbf{0.938} & \textbf{0.982} & \textbf{0.839} \\
        \hline\hline
         \multirow{2}{*}{Methods} & \multicolumn{2}{c|}{\textit{OrganMNIST3D}} & \multicolumn{2}{c|}{\textit{NoduleMNIST3D}} & \multicolumn{2}{c|}{\textit{FractureMNIST3D}} & \multicolumn{2}{c|}{\textit{AdrenalMNIST3D}} & \multicolumn{2}{c|}{\textit{VesselMNIST3D}} & \multicolumn{2}{c}{\textit{SynapseMNIST3D}} \\
         \cline{2-13}
          & AUC & ACC & AUC & ACC & AUC & ACC & AUC & ACC & AUC & ACC & AUC & ACC \\
          \hline
        ResNet-18 (2.5D) & 0.977 & 0.788 & 0.838 & 0.835 & 0.587 & 0.451 & 0.718 & 0.772 & 0.748 & 0.846 & 0.634 & 0.696 \\
        ResNet-18 (3D) & 0.996 & 0.907 & 0.863 & 0.844 & 0.712 & 0.508 & 0.827 & 0.721 & 0.874 & 0.877 & 0.820 & 0.745 \\
        ResNet-18 (ACS) & 0.994 & 0.900 & 0.873 & 0.847 & 0.714 & 0.497 & 0.839 & 0.754 & 0.930 & 0.928 & 0.705 & 0.722 \\
        ResNet-50 (2.5D) & 0.974 & 0.769 & 0.835 & 0.848 & 0.552 & 0.397 & 0.732 & 0.763 & 0.751 & 0.877 & 0.669 & 0.735 \\
        ResNet-50 (3D) & 0.994 & 0.883 & 0.875 & 0.847 & 0.725 & 0.494 & 0.828 & 0.745 & 0.907 & 0.918 & 0.851 & 0.795 \\
        ResNet-50 (ACS) & 0.994 & 0.889 & 0.886 & 0.841 & 0.750 & 0.517 & 0.828 & 0.758 & 0.912 & 0.858 & 0.719 & 0.709 \\
        auto-sklearn & 0.977 & 0.814 & 0.914 & 0.874 & 0.628 & 0.453 & 0.828 & 0.802 & 0.910 & 0.915 & 0.631 & 0.730 \\
        AutoKeras & 0.979 & 0.804 & 0.844 & 0.834 & 0.642 & 0.458 & 0.804 & 0.705 & 0.773 & 0.894 & 0.538 & 0.724 \\
        Ours & \textbf{0.999} & \textbf{0.966} & \textbf{0.938} & \textbf{0.887} & \textbf{0.760} & \textbf{0.608} & \textbf{0.878} & \textbf{0.829} & \textbf{0.953} & \textbf{0.935} & \textbf{0.906} & \textbf{0.881} \\  
    \end{tabular}}
    \caption{Comparison between our model and benchmark in all MedMNIST datasets. We highlight the best ACC and best AUC for each dataset.}
    \label{tab:benchmark_mnist}
\end{table}

Additionally, we perform the same experiments for datasets with more realistic image sizes. In Table~\ref{tab:finetune_real}, we show the impact of the choice of the number of stages to extract local features in the model. 
One stage means features are extracted from the Stage 4. Two stages is represented in Figure~\ref{fig:layout}. Three stages means we also include local features from the Stage 2. 
In the case of ISIC2018, the same behavior pattern existing in MedMNIST is observed for pre-trained backbone, where two or three stages yields similar results.
However, when fine-tuning is applied, any choice of number of stages show similar results.
In the case of COVID-CT, the use of two stages for feature extration is optimal for both pre-trained and fine-tuned backbones.
This indicates that, using images that contain more information than the small sized ones from MedMNIST(v2), relevant features is also captured by earlier layers. In those cases, fine-tuning is not detrimental as the amount of information in the datasets is enough for overfitting not to occur.

\begin{table}
    \centering
    \caption{Accuracy and Area Under ROC Curve of the model with pre-trained and fine-tuned backbone. We show the results for varying number of stages used for feature extraction. Stages are chosen from later to earlier ones.}
    \label{tab:finetune_real}
    \begin{tabular}{ l c | c c | c c }
    \multirow{2}{*}{Datasets} & \multirow{2}{*}{Stages} & \multicolumn{2}{c|}{\textit{Pre-trained backbone}} & \multicolumn{2}{c}{\textit{Fine-tuned backbone}} \\
    \cline{3-6}
     &  & ACC & AUC & ACC & AUC \\
     \hline
     COVID-CT & 1 & 0.768 & 0.856 & 0.803 & 0.870 \\ 
     COVID-CT & 2 & 0.788 & 0.868 & 0.813 & 0.866 \\ 
     COVID-CT & 3 & 0.773 & 0.864 & 0.805 & 0.855 \\ 
     \hline
     ISIC2018 & 1 & 0.845 & 0.963 & 0.863 & 0.969 \\ 
     ISIC2018 & 2 & 0.860 & 0.971 & 0.862 & 0.967 \\
     ISIC2018 & 3 & 0.865 & 0.972 & 0.862 & 0.971 \\ 
    \end{tabular}
\end{table}

Finally, in Table~\ref{tab:benchmark_real} we compare our model with recent literature. In the cases where results for the datasets used here are not found in liteture, we proceed with the same setup we used for fine-tuning the MIT-EfficientViT backbone of our model. The choice of models was based on the number of parameters, where we limited it to 25 million. Our model outperforms all cited literature in the case of Clean-CC-CCII and most in the case of ISIC2018. This, combined with the results in MedMNIST(v2), indicates our model is suitable for biomedical applications where the amount of data available for training is limited.

\begin{table}
    \centering
    \caption{Comparison between our model and other methods in the literature in all MedMNIST3D datasets. We highlight the best ACC and best AUC for each dataset.}
    \label{tab:benchmark_real}
    \begin{tabular}{| l | c c | c c |}
        \hline
        \multirow{2}{*}{Methods} & \multicolumn{2}{c|}{\textit{Clean-CC-CCII}} & \multicolumn{2}{c|}{\textit{ISIC2018}}  \\
        \cline{2-5}
         & AUC & ACC & AUC & ACC \\
        \hline
        ResGANet-101~\cite{cheng2022resganet} & 0.780 & 0.820 & - & 0.824 \\
        Hiera-T~\cite{ryali2023hiera} & 0.635 & 0.581 & 0.881 & 0.719 \\
        MogaNet-T~\cite{li2024moganet} & 0.783 & 0.853 & 0.945 & 0.804 \\
        MNv4-Hybrid-M~\cite{qin2025mobilenetv4} & 0.730 & 0.675 & 0.842 & 0.698 \\
        MambaOut-T~\cite{yu2024mambaout} & 0.788 & 0.863 & 0.973 & 0.870 \\
        Ours & 0.813 & 0.866 & 0.967 & 0.862 \\  
        \hline
    \end{tabular}
\end{table}

\section{Conclusions}
\label{sec:conclusions}

In this work, we developed an approach to orderlessly encode information extracted from multiple stages of a Vision Transformer model. 
More specifically, we extracted local features from one to three stages of MIT-EfficientViT and developed a lossless method to concatenate those features. 
We developed a methodology to sample the training dataset and used this sample to compute the concatenated set of local features. In this way, we were able to estimate the GMM for large datasets using few computational resources.
We developed a Fisher Vector layer to perform the orderless enconding, which was used to replace the global pooling approach present in MIT-EfficientViT \cite{cai2023efficientvit}. The proposed approach was evaluated in all MedMNIST \cite{yang2023medmnist} datasets, where it outperformed benchmark in both 2D and 3D datasets. 

Additionally, this work proposes an approach for increasing the model's accuracy by improving local features explanation capacity on target datasets. This is performed by fine-tuning the backbone prior to the GMM estimation. In the particular case of MedMNIST3D, where no pretrained weights are available, we preload the backbone with pretrained 2D weights. Fine-tuning proved to be beneficial, on average, in both 2D and 3D cases.

We further explore the performance of the model by comparing it to recent literature in datasets where image sizes are closer to realistic applications. We used Clean-CC-CCII and ISIC2018 for this task and obtained competitive results with literature maintaining similar model sizes in terms of parameter count.

However, given the fact that fine-tuning is performed before the GMM estimation, the obtained features may be suboptimal for Fisher Vector encoding. In future works, we intend to investigate alternative methods for first- and second-order statistics estimation. We are particularly interested in methods that are trainable through backpropagation algorithm. This enables end-to-end learning and may result in optimal local features. Furthermore, methods that allow incremental learning remove the necessity of sampling the training set and may obtain optimal first- and second-order statistics.

\section*{Code availability}

The code developed in this study is publicly available at \url{https://github.com/lolyra/medical}.

\section*{Declaration of competing interest}

J. B. F. reports equipment, drugs, or supplies was provided by State of Sao Paulo Research Foundation. J. B. F. reports financial support was provided by National Council for Scientific and Technological Development. L. O. L. reports financial support was provided by Coordination of Higher Education Personnel Improvement.

\section*{Acknowledgements}

This study was financed in part by the Coordenação de Aperfeiçoamento de Pessoal de Nível Superior - Brasil (CAPES) - Finance Code 001.
J. B. F. gratefully acknowledges the financial support of S\~ao Paulo Research Foundation (FAPESP) (Grants \#2020/01984-8 and \#2020/09838-0) and of 
National Council for Scientific and Technological Development, Brazil (CNPq) (Grant \#306981/2022-0).


\end{document}